\documentclass{article}
\usepackage{spconf,amsmath,epsfig}
\usepackage{amsfonts}
\usepackage{upgreek}
\usepackage{graphicx}
\usepackage{makecell}
\usepackage{array}
\usepackage{units}
\usepackage[square,numbers]{natbib}
\usepackage{caption, setspace}
\usepackage[activate={true,nocompatibility},final,tracking=true,kerning=true,spacing=true,factor=1100,stretch=10,shrink=10]{microtype}
\usepackage[hidelinks]{hyperref}

\title{A Novel Multi-Attention Driven System for Multi-Label Remote Sensing Image Classification}

\name{Gencer Sumbul and Beg\"{u}m Demir}
\address{Faculty of Electrical Engineering and Computer Science, Technische Universit\"at Berlin, Germany}

\begin{document}
%\ninept
%
\maketitle
\begin{abstract}
  This paper presents a novel multi-attention driven system that jointly exploits Convolutional Neural Network (CNN) and 
  Recurrent Neural Network (RNN) in the context of 
  multi-label remote sensing (RS) image classification. 
  The proposed system consists of four main modules. 
  The first module aims to extract preliminary local descriptors of RS image bands that can be associated to different spatial resolutions. 
  To this end, we introduce a K-Branch CNN, in which each branch extracts descriptors of image bands that have the same spatial resolution. The second module aims to model spatial relationship among local descriptors. This is achieved by a bidirectional RNN architecture, in which Long Short-Term Memory nodes enrich local descriptors by considering spatial 
  relationships of local areas (image patches). The third module aims to define multiple attention scores for local 
  descriptors. This is achieved by a novel patch-based 
  multi-attention mechanism that takes into account the joint occurrence of multiple land-cover classes and provides the attention-based local descriptors. 
  The last module exploits these descriptors for multi-label RS image 
  classification. Experimental results obtained on the BigEarthNet that is a large-scale Sentinel-2 benchmark archive 
show the effectiveness of the proposed method compared to a state of the art method. 
\end{abstract}

\vspace{-0.07in}
\section{Introduction}
\vspace{-0.07in}
\label{sec:intro}

The increased number of recent Earth observation satellite missions has led to a significant growth of remote 
sensing (RS) image archives. Accordingly, classification of RS image scenes, 
which is usually achieved by direct supervised classification of each image in the archive, 
has received increasing attention in RS~\cite{Dai:2018, Cheng:2017}. 
Most of the supervised classifiers developed for RS image scene classification 
problems are trained on images annotated with single 
high-level category labels, which are associated to the most significant content of the image. 
However, each RS image typically consists of multiple classes and thus can be simultaneously associated to different low-level land-cover class labels 
(i.e., multi-labels)~\cite{Dai:2018}. 
Thus, supervised classification methods that exploit training images annotated by 
multi-labels have been recently introduced in RS. 
As an example, a sparse reconstruction-based RS image classification and 
retrieval method is introduced in~\cite{Dai:2018}. This method considers a measure of label likelihood in the 
framework of sparse reconstruction-based classifiers and generalizes the original sparse 
classifier to the case of multi-label RS image classification and retrieval problems. 
Recent advances in deep learning have attracted great attention in RS image classification 
problems. In~\cite{Hua:2017}, a Recurrent Neural Network (RNN) architecture that sequentially 
models the co-occurrence relationships of land-cover classes with class-specific 
attention extracted from a Convolutional Neural Network (CNN) is introduced in the 
framework of multi-label image scene classification. In this method, inaccurate prediction of any land-cover class label can be propagated through the RNN sequence and this may affect the accurate prediction of other classes in the same image. 
This issue is known as error propagation problem~\cite{Senge:2014}. To address this problem, we propose a system that includes a novel patch-based multi-attention mechanism. The proposed mechanism allows accurate characterization of 
the joint occurrence of different land-cover classes and thus reduces the effect of the error propagation problem. 
\vspace{-0.07in}
\section{Proposed Multi-Attention Driven System}
\begin{figure}[t]
  \centering
  \captionsetup{justification=justified,singlelinecheck=false, format=hang}
  \includegraphics[trim=0 10 0 0,clip,width=\linewidth]{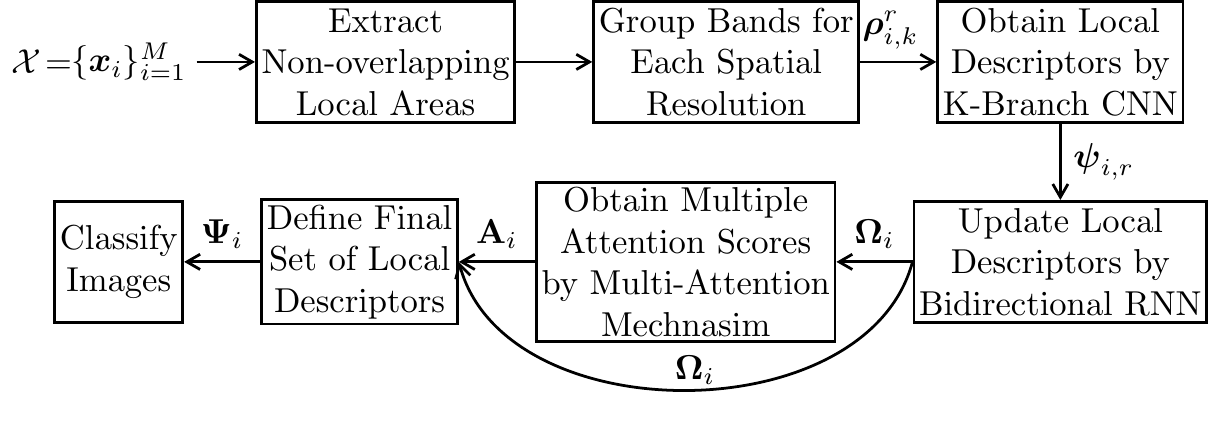}
  \caption{Block diagram of the proposed system with a novel patch-based multi-attention mechanism.}
  \label{fig:framework}
  \vspace{-0.3cm}
\end{figure}

\vspace{-0.07in}
\label{sec:method}
Let \mbox{$\mathcal{X}{=}\{\boldsymbol{x}_1,...,\boldsymbol{x}_M{\}}$} be an archive that consists of $M$ images, where $\boldsymbol{x}_i$ is the $i$\textsuperscript{th} image. 
We assume that each image in $\mathcal{X}$ is associated with multi-labels from a label set 
\mbox{$\mathcal{L} = \{l_1,...,l_C\}$} with $|\mathcal{L}| = C$. Label information of $\boldsymbol{x}_i$ is defined by 
a binary vector $\boldsymbol{y}_i \in \{0,1\}^{C}$, where each element of 
$\boldsymbol{y}_i$ indicates the presence or absence of 
label $l_c \in \mathcal{L}$ in a sequence. We assume that for a given image $\boldsymbol{x}_i$, 
the spectral bands can be associated to the 
$K$ different spatial resolutions, resulting in different pixel sizes. 
We aim to learn $F(\boldsymbol{x}^*)=g(f(\boldsymbol{x}^*))$ that maps a new image 
$\boldsymbol{x}^*$ to multi-labels, where $f(\cdot)$ generates classification 
scores for each label of $\mathcal{L}$ and $g(\cdot)$ produces $\boldsymbol{y}^*$ as a 
predicted label set. The proposed system is characterized by four
main modules: 1) extraction of local descriptors by a K-Branch CNN;
2) characterization of spatial relationships among local areas (image patches) 
with an LSTM based bidirectional RNN; 
3) defining multiple attention scores for local descriptors by a novel patch-based multi-attention mechanism; 
and 4) multi-label scene classification. 
Fig.~\ref{fig:framework} presents the block diagram of the proposed system and each module is explained in the 
following sections. 
\vspace{-0.1in}
\subsection{Extraction of Local Descriptors by the Proposed K-Branch CNN}
\vspace{-0.05in}
\label{ssec:cnn}
This module aims to extract local descriptors of each image in the archive. 
We assume that each image band can be associated to different spatial resolutions (e.g., Sentinel-2 image bands). 
To this end, we introduce $K$-Branch CNN to use different CNNs specialized for RS image 
bands with different spatial resolutions. We would like to 
note that if all bands are associated with the same spatial resolution, 
this module becomes a single branch CNN (i.e., $K=1$). 
We initially divide each image to $R$ non-overlapping local areas (i.e., patches) and then define 
different sets of bands based on their spatial resolution. Let $\boldsymbol{\rho}_{i,k}^r$ be the 
$k$\textsuperscript{th} subset of the $r$\textsuperscript{th} patch 
with a given spatial resolution, where $k \in \{1,2,...,K\}$ and 
$r \in \{1,2,...,R\}$. Let $\phi^k$ be the $k$\textsuperscript{th} branch that generates a descriptor for 
$\boldsymbol{\rho}_{i,k}^r$ by applying convolutional layers and one fully connected (FC) 
layer. After obtaining different local descriptors for all subsets, 
they are concatenated into one vector for one patch. Then, a 
new FC layer takes all concatenated vectors and 
produces the local descriptors $\boldsymbol{\psi}_{i,r}$. 
Our K-Branch CNN module is illustrated in Fig.~\ref{fig:KB_CNN}.
\vspace{-0.11in}
\subsection{Characterization of Spatial Relationship among Image Patches}
\begin{figure}[t]
  \centering
  \captionsetup{justification=justified,singlelinecheck=false, format=hang}
  \includegraphics[width=\linewidth, height=0.65\linewidth]{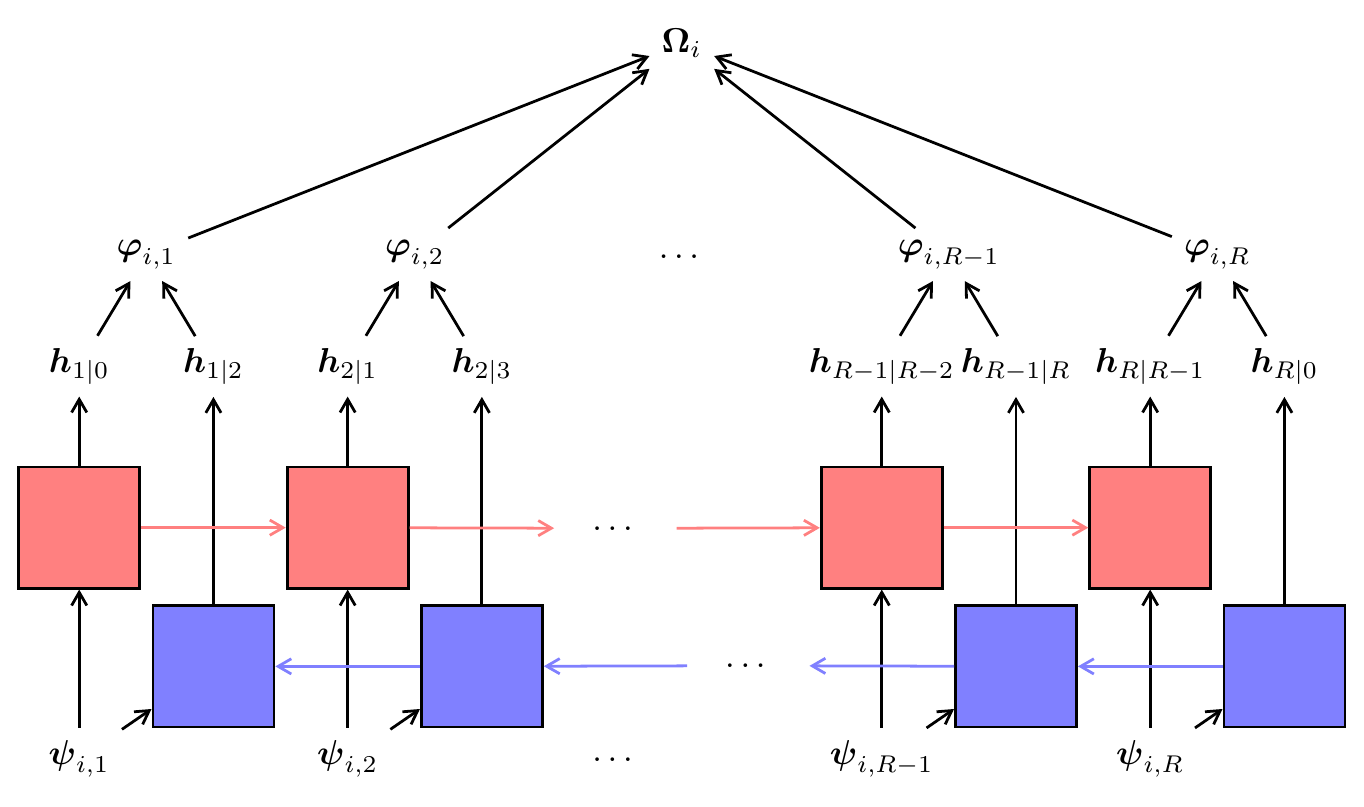}

  \caption{The proposed bidirectional RNN module. Forward and backward passes are shown in red and blue, respectively.}
  \label{fig:RNN}
  \vspace{-0.4cm}
\end{figure}

\vspace{-0.05in}
\label{sec:rnn}
This module aims to: i) model the spatial relationships of image patches; 
and ii) update the local descriptors based on this information. To this end, 
we utilize two RNNs in a bidirectional manner. 
Each node of the first RNN updates the descriptor of one patch concerning previous patches 
(i.e., previous nodes). The second RNN employs the same idea by considering the subsequent patches. 
For the nodes of each RNN, we use the LSTM~\cite{lstm} neurons 
because of their capacity to capture long-term dependencies among nodes. Thus, each LSTM node 
takes the descriptor of the $r$\textsuperscript{th} patch ($\boldsymbol{\psi}_{i,r}$) from the K-Branch CNN 
as input and updates this descriptor concerning the successive local descriptors as follows:
\begin{figure}[t]
  \centering
  \captionsetup{justification=raggedright,singlelinecheck=false, format=hang}
  \includegraphics[width=\linewidth]{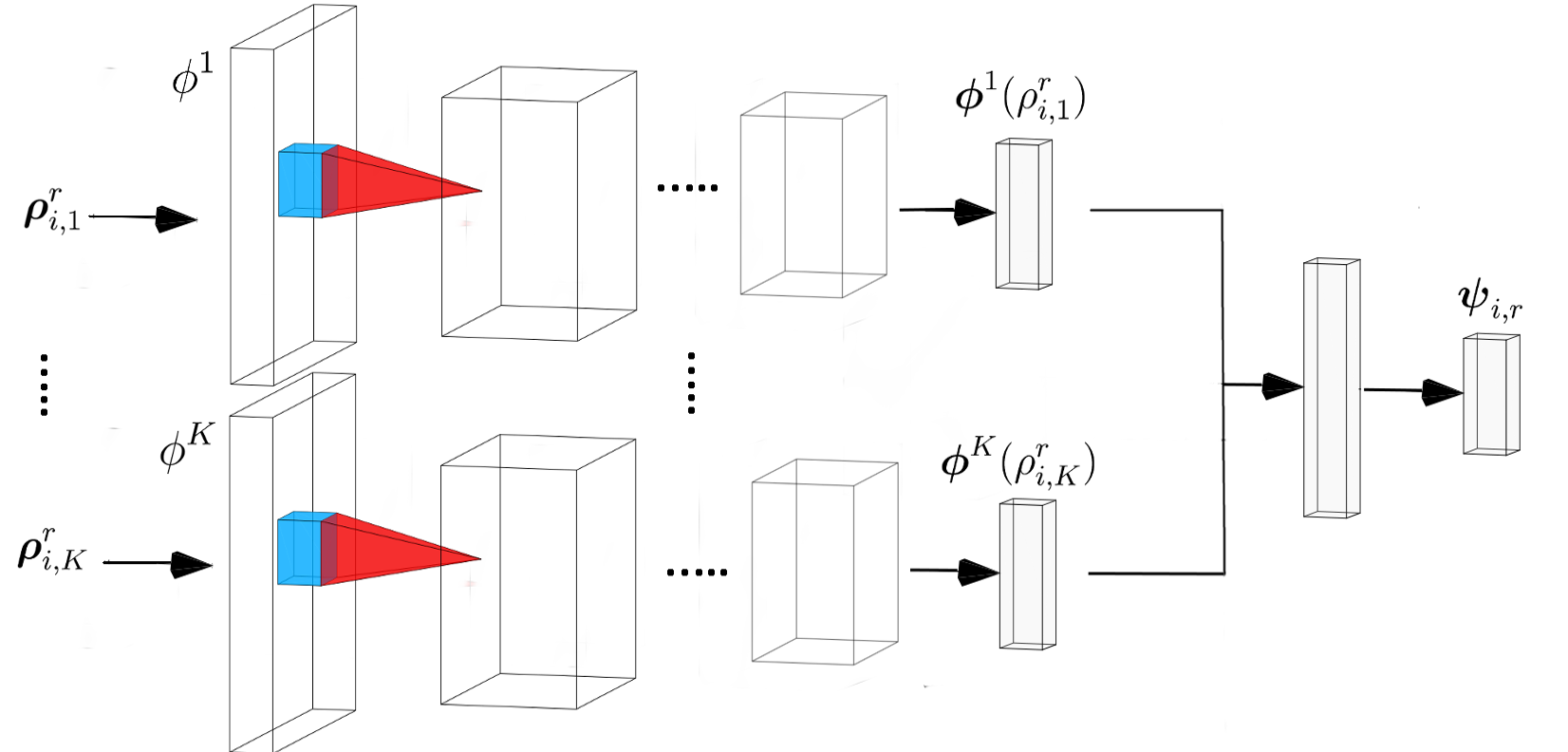}
  \vspace{0.05cm}
  \caption{The proposed K-Branch CNN to obtain local descriptors.}
  \label{fig:KB_CNN}
  \vspace{-0.2cm}
\end{figure}
\begin{equation}
  \begin{aligned}
    \boldsymbol{f}_{r}  &= \! \delta(\mathbf{W}_{f,r}\boldsymbol{\psi}_{i,r} + \mathbf{U}_{f,r}\boldsymbol{h}_{\uptau} + \boldsymbol{b}_{f,r}) \\
    \boldsymbol{i}_{r}  &= \! \delta(\mathbf{W}_{i,r}\boldsymbol{\psi}_{i,r} + \mathbf{U}_{i,r}\boldsymbol{h}_{\uptau} + \boldsymbol{b}_{i,r}) \\
    \boldsymbol{o}_{r}  &=  \! \delta(\mathbf{W}_{o,r}\boldsymbol{\psi}_{i,r} + \mathbf{U}_{o,r}\boldsymbol{h}_{\uptau} + \boldsymbol{b}_{o,r}) \\
    \boldsymbol{c}_{r}  &= \!\! \boldsymbol{f}_{r}\odot \boldsymbol{c}_{\uptau} \! +  \boldsymbol{i}_{r}\odot\tanh(\mathbf{W}_{c,r}\boldsymbol{\psi}_{i,r} \! + \mathbf{U}_{c,r}\boldsymbol{h}_{\uptau} \! + \boldsymbol{b}_{c,r}) \\
    \boldsymbol{h}_{r}  &= \! \boldsymbol{h}_{r|\uptau} = \boldsymbol{o}_{r} \odot \tanh(\boldsymbol{c}_{r})
  \end{aligned}
\end{equation}
where $\tanh$ is the hyperbolic tangent function, $\delta$ is the sigmoid function, 
$\mathbf{W}_{\boldsymbol{.},r}$ and $\boldsymbol{b}_{\boldsymbol{.},r}$ are the weight and 
bias parameters. 
$\boldsymbol{i}$, $\boldsymbol{f}$, $\boldsymbol{o}$ and $\boldsymbol{c}$ are input gate, 
forget gate, output gate and cell state, respectively (for a detailed explanation, see~\cite{Gers:2000}). 
For bidirectional RNN architecture, 
we use two LSTM nodes (for forward and backward passes) related to each image patch 
with different parameters. For the first and second time order, $\uptau$ becomes 
$r-1$ and $r+1$, respectively since two RNNs update local descriptors in the reversed order. 
For the beginning of passes ($r=1$ or $r=R$), $\uptau$ refers to an initial state of the nodes. 
Thus, we obtain the sequentially updated $r$\textsuperscript{th} local descriptor of $\boldsymbol{x}_i$ by concatenating the outputs of the 
two LSTM nodes as follows:
\begin{equation}
  \boldsymbol{\varphi}_{i,r} = [\boldsymbol{h}_{r|r-1}^{\top}, \boldsymbol{h}_{r|r+1}^{\top}]^{\top}.
\end{equation}
Our bidirectional RNN module is illustrated in Fig.~\ref{fig:RNN}.
\vspace{-0.1in}
\subsection{Defining Multiple Attention Scores for Image Patches by a Novel Multi-Attention Mechanism}
\vspace{-0.05in}
\label{sec:attention}
Each local descriptor can be associated with multiple land-cover classes. This 
module aims to determine the attention scores for each local descriptor. Attention scores show 
the relevancy of considered image patch in terms of the land-cover classes present in the whole image for the 
complete characterization of its semantic content. To this end, we introduce a novel multi-attention mechanism to obtain 
multiple attention scores as follows:
\setlength{\abovedisplayskip}{5pt}
\setlength{\belowdisplayskip}{5pt}
\begin{equation}
  \mathbf{A}_i = \sigma(\mathbf{W}_{a,2}\tanh(\mathbf{W}_{a,1}\mathbf{\Omega}_{i})) 
\end{equation}
where $\mathbf{\Omega}_{i}$ denotes a matrix, whose columns contain local descriptors. 
$\mathbf{W}_{a,1}$ and $\mathbf{W}_{a,2}$ are the weight parameters of the attention mechanism and
$\sigma$ is the softmax function that produces normalized weights within the interval $[0,1]$. 
This attention mechanism can be regarded as two FC layers that use hyperbolic tangent and softmax activation 
functions without bias parameters. Therefore, attention matrix $\mathbf{A}_i \in [0,1]^{T \times R}$ provides $T$ different attention scores for each patch. 
Then, the final set $\mathbf{\Psi}_{i}$ of descriptors for $\boldsymbol{x}_i$ can be obtained by: i) multiplication of 
$\mathbf{\Omega}_{i}$ with the attention matrix; and ii) addition of non-linearity with $\max$ function 
to use $\mathbf{\Psi}_{i}$ for the multi-label classification module. Attending to each patch 
with multiple scores allows us to consider the joint occurrence of all land-cover classes in the label sequence. 
For instance, an increase in one of the attention scores of a local descriptor for \textit{beach}
class will increase the attention score of the same local descriptor for the \textit{sea} class while 
it may not be the case for other local descriptors. 

It is worth noting that a single attention score could be used for this module. However, since multi-label image classification problem requires to jointly learn multiple classes, the use of single attention score is not sufficient and multiple attention scores are necessary.
\begin{figure}
  \captionsetup{justification=raggedright,singlelinecheck=false, format=hang, font=scriptsize}
  \begin{minipage}[c]{0.17\linewidth}
    \includegraphics[width=\columnwidth]{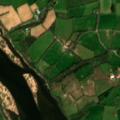}
  \end{minipage}
  \hfill
  \begin{minipage}[c]{0.3\linewidth}
    \captionsetup{justification=raggedright,singlelinecheck=false, format=hang, font=scriptsize}
    \caption*{non-irrigated arable land, pastures, land principally occupied by agriculture, inland marshes, water courses}
  \end{minipage}
  \hfill
  \begin{minipage}[c]{0.17\linewidth}
    \includegraphics[width=\columnwidth]{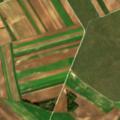}
  \end{minipage}
  \hfill
  \begin{minipage}[c]{0.3\linewidth}
    \captionsetup{justification=raggedright,singlelinecheck=false, format=hang, font=scriptsize}
    \caption*{non-irrigated arable land, vineyards, pastures, land principally occupied by agriculture}
  \end{minipage}
  \hfill
  \vspace{-0.1cm}
  \begin{minipage}[c]{0.17\linewidth}
    \includegraphics[width=\columnwidth]{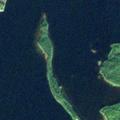}
  \end{minipage}
  \hfill
  \begin{minipage}[c]{0.3\linewidth}
    \captionsetup{justification=raggedright,singlelinecheck=false, format=hang, font=scriptsize}
    \caption*{coniferous forest, water bodies}
  \end{minipage}
  \hfill
  \begin{minipage}[c]{0.17\linewidth}
    \includegraphics[width=\columnwidth]{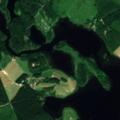}
  \end{minipage}
  \hfill
  \begin{minipage}[c]{0.3\linewidth}
    \captionsetup{justification=raggedright,singlelinecheck=false, format=hang, font=scriptsize}
    \caption*{non-irrigated arable land, land principally occupied by agriculture, coniferous forest, mixed forest, water bodies}
  \end{minipage}
  \hfill
  \vspace{-0.23 cm}
  \begin{minipage}[c]{0.17\linewidth}
    \includegraphics[width=\columnwidth]{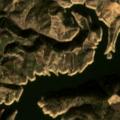}
  \end{minipage}
  \hfill
  \begin{minipage}[c]{0.3\linewidth}
    \captionsetup{justification=raggedright,singlelinecheck=false, format=hang, font=scriptsize}
    \caption*{olive groves, land principally occupied by agriculture, broad-leaved forest, transitional woodland/shrub, water bodies}
  \end{minipage}
  \hfill
  \begin{minipage}[c]{0.17\linewidth}
    \includegraphics[width=\columnwidth]{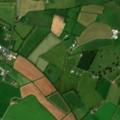}
  \end{minipage}
  \hfill
  \begin{minipage}[c]{0.3\linewidth}
    \captionsetup{justification=raggedright,singlelinecheck=false, format=hang, font=scriptsize}
    \caption*{non-irrigated arable land, pastures, complex cultivation patterns}
  \end{minipage}
  \hfill
  \captionsetup{justification=raggedright,singlelinecheck=false, format=hang, font={small,stretch=0.75}}
  \caption{Example of Sentinel-2 images and their multi-labels in the BigEarthNet archive.}
  \label{fig:patch_ex}
  \vspace{-0.3cm}
\end{figure}
\vspace{-0.1in}
\subsection{Classification of RS Image Scenes with Multi-Labels}
\vspace{-0.05in}
\label{sec:classification}
This module aims to assign multi-labels to RS images by using the final set $\mathbf{\Psi}_{i}$ of descriptors. 
To this end, a classification layer $f(\cdot)$ that generates class scores $z_{l_j}$ 
for each class label $l_j$ in a sequence by using the vectorized $\mathbf{\Psi}_{i}$ is considered. 
Finally, we obtain the class posterior probability of $l_j$ for the image $\boldsymbol{x}_i$ 
with the \textit{sigmoid} function as: $P(l_j|\boldsymbol{x}_i) = \nicefrac{1}{(1+e^{-z_{l_j}})}$. 
After end-to-end training of the whole neural network by minimizing cross-entropy loss, 
the proposed system predicts the multi-labels of a new image $\boldsymbol{x}^*$ by thresholding the probability 
values. 

We would like to note that unlike~\cite{Hua:2017}, our system predicts multi-labels of an RS image by considering all outputs of an RNN instead of deciding each label by considering a single class specific node.
\vspace{-0.07in}
\section{Experimental Results}
\vspace{-0.07in}
\label{sec:exp}
Experiments were conducted on the BigEarthNet benchmark archive\footnote{The BigEarthNet is available at \url{http://bigearth.net}.} that was constructed by selecting 
125 Sentinel-2 tiles distributed over 10 European countries (Austria, Belgium, Finland, 
Ireland, Kosovo, Lithuania, Luxembourg, Portugal, Serbia, Switzerland) and acquired between 
June 2017 and May 2018~\cite{bigearthnet}. 
In details, all the considered image tiles are associated with cloud cover 
percentage less than $1\%$ and were atmospherically corrected by the Sentinel-2 Level 2A product generation and formatting tool (sen2cor). All spectral bands except the 10\textsuperscript{th} band, 
for which surface information is not embodied, were included. The tiles were divided into $590,326$ 
non-overlapping images. Each image in the archive is a section of: i) $120\times120$ pixels for 10m bands; ii) $60\times60$ pixels for 20m bands; and iii) $20\times20$ pixels for 60m bands. 
Each image in the archive has been annotated with one or more land-cover class labels (i.e., multi-labels) provided from the CORINE Land Cover (CLC) 
database of the year 2018 (CLC 2018). We included Level-3 CLC classes except 
glaciers and perpetual snow class (and thus $43$ land-cover classes are included). 
The number of labels associated to each image 
varies between $1$ and $12$, whereas $95\%$ of images have at most $5$ multi-labels. 
Fig.~\ref{fig:patch_ex} shows an example of images and the multi-labels associated with them. 
The BigEarthNet will make a significant advancement in terms of developments of algorithms for the analysis of large-scale RS image archives.

In the experiments, $70,987$ images that are fully covered by seasonal snow, cloud and cloud shadow were eliminated\footnote{The lists of images fully covered by seasonal snow, cloud and cloud shadow are available at \url{http://bigearth.net/\#downloads}.}. Then, among the remaining images, we randomly selected: i) $60\%$ of images to derive a training set; ii) $20\%$ of images to derive a validation set; and iii) $20\%$ of images to derive a test set. Then, we divided each image into $16$ non-overlapping patches. Since Sentinel-2 image bands are associated to three different spatial resolutions, we selected $K=3$ for the K-Branch CNN. Thus, we split the bands into three subsets for each patch and in each subset we stacked 
bands into a single volume for its corresponding CNN branch. 
The first branch takes as input the bands $2$ to $4$ and $8$ (which have $10$m spatial resolution), while the second branch takes as input the bands $5$ to $7$, $8$A, $11$ and $12$ (which have $20$m spatial resolution) 
and the third branch takes as input the bands $1$ and $9$ (which have $60$m spatial resolution). 
Selection of the number of patches and all other hyperparameters 	
was achieved based on the classification performance on the validation set. All branches have the same regime for the number of filters of convolutional layers. 
We used $32$ filters for the initial layers. While adding new layers, the number of filters is 
first increased to the multiplication of $2$ and then decreased to the division of $2$. It ends with $64$ filters for all branches. However, filter size and applied operations between convolutional 
layers of each branch are different. $5\times5$ filters for initial layers and $3\times3$ filters for deeper layers are used for the first branch. 
For the second and third branches, $3\times3$ filters and $2\times2$ filters are used throughout the layers, respectively. For all convolutional layers, stride of 1 and zero padding were used. 
In addition, we applied max-pooling for the first two branches. 
However, in order not to decrease spatial resolution more, it was not applied to the last branch.  
For each branch, an FC layer, which takes the output of the last convolutional layer and produces spatial 
resolution specific description of an image portion, was considered. For the bidirectional RNN module, we exploited the LSTM nodes with $128$-dimensional hidden units. 
\begin{table}[t]
  \setlength{\tabcolsep}{7pt}
  \captionsetup{justification=justified,singlelinecheck=false, format=hang}
  \caption{Results obtained by the CA-ConvBiLSTM and our system.}
  \centering
  \label{result}
  \begin{tabular}{rccc}
  \hline %\noalign{\smallskip} \hline \noalign{\smallskip}
  Method & $R$ ($\%$) & $F_1$ & $F_2$\\
  \hline
  CA-ConvBiLSTM~\cite{Hua:2017} & $77.56$ & $0.7089$ & $0.7405$ \\
  Proposed System & $80.0$ & $0.7289$ & $0.7631$ \\
  \hline
  \end{tabular}
  \vspace{-0.3cm}
\end{table}

\begin{figure}[t]
  \centering
  \captionsetup{justification=justified,singlelinecheck=false, format=hang}
  \includegraphics[width=\linewidth]{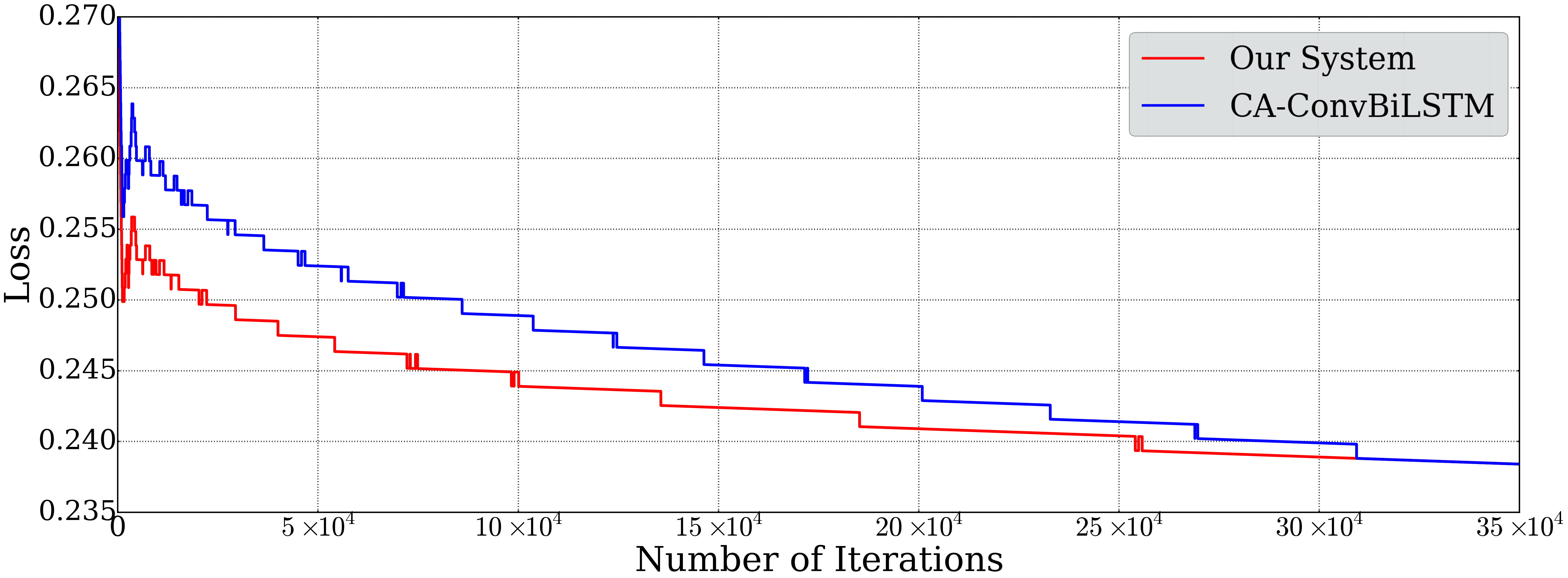}
  \caption{Training trajectory of the loss for our system and the CA-ConvBiLSTM. To display better, we used moving averages.}
  \label{fig:loss}
  \vspace{-0.3cm}
\end{figure}

In the experiments, we compared our system with the CA-ConvBiLSTM~\cite{Hua:2017}, 
which is the only multi-label RS image classification method that considers deep learning. 
For a fair comparison, we used our first branch as the feature 
extraction module of the CA-ConvBiLSTM. We applied cubic interpolation to $20$m and $60$m bands 
in order to stack all bands into one volume, which was fed into the first branch. 
For the RNN part of the CA-ConvBiLSTM, we used the same type of LSTM nodes as in our system. 
We applied the same end-to-end training procedure from scratch for all experiments 
to compare both methods under the same setting. 
All model parameters were initialized with Xavier method~\cite{Glorot:2010} except the suggested 
LSTM initialization of the CA-ConvBiLSTM. We trained models $100$ epochs with the initial 
learning rate of $10^{-3}$. 
Results of each method are provided in terms of three performance evaluation metrics: 1) Recall ($R$), 2) $F_1$ Score and 
3) $F_2$ Score~\cite{Zhang:2014}. As we can see from Table~\ref{result}, our system provides $2.44\%$ higher recall, $0.02$ higher $F_1$ score and $0.0226$ higher $F_2$ score compared to the CA-ConvBiLSTM. The performance 
improvements on all metrics are statistically significant under a value of $p \ll 0.0001$. 
In terms of training efficiency, our system achieves the lowest training loss 
earlier than the CA-ConvBiLSTM (see Fig.~\ref{fig:loss}). This is very important for large-scale benchmark archives 
like BigEarthNet. We would like to note that the promising performance of our system compared to the CA-ConvBiLSTM relies on the efficient characterization of: 
i) local descriptors of image bands with different spatial resolutions; and ii) multiple attention scores for 
each local descriptor that evaluate the joint occurrence of multiple land-cover 
classes.  
\vspace{-0.07in}
\section{Conclusion}
\vspace{-0.07in}
\label{sec:conc}
This paper proposes a novel multi-attention driven system for classification of 
RS images with multi-labels. The proposed system consists of four main modules: 1) the spatial resolution specific K-Branch CNN for the extraction of local descriptors; 2) the bidirectional RNN architecture with LSTM nodes for modeling the spatial relationships of the image patches; 3) the patch-based 
multi-attention mechanism to obtain multiple attention scores for each local descriptor; and 
4) multi-label classification of images in the archive. 
Experimental results obtained on the BigEarthNet archive 
show the effectiveness of our system. As a future work, we plan to extract and exploit image regions defined by segmentation algorithms 
for the characterization of local descriptors.
\small
\vspace{-0.03in}
\section{Acknowledgements}
\vspace{-0.05in}
This work was supported by the European Research Council under the ERC Starting Grant BigEarth-759764.
%\vspace{-0.07in}
\bibliographystyle{IEEEbib}
%\setstretch{0.87}
\bibliography{defs,refs}

\end{document}